\documentclass[11pt]{article}
\usepackage{eacl2017}
\usepackage{times}
\usepackage{url}
\usepackage{latexsym}
\usepackage{amsmath}
\usepackage{amssymb}
\usepackage{breqn}
\usepackage{booktabs}
\usepackage{pbox}
\usepackage{multirow}
\usepackage{graphicx}
\usepackage[round]{natbib}
\usepackage{tablefootnote}

\eaclfinalcopy 
\def\eaclpaperid{501} 

\title{Latent Variable Dialogue Models and their Diversity}

\author{Kris Cao \and Stephen Clark \\
  Computer Laboratory \\
  University of Cambridge \\
  United Kingdom \\
  {\tt \{kc391, sc609\}@cam.ac.uk} \\}

\date{}

\begin{document}
\maketitle
\begin{abstract}
  We present a dialogue generation model that directly captures the variability in possible responses to a given input, which reduces the `boring output' issue of deterministic dialogue models. Experiments show that our model generates more diverse outputs than baseline models, and also generates more consistently acceptable output than sampling from a deterministic encoder-decoder model.
\end{abstract}

\section{Introduction}

The task of open-domain dialogue generation is an area of active development, with neural sequence-to-sequence models dominating the recently published literature \citep{Shang:15,Vinyals:15,Jiwei:16b,Jiwei:16,Serban:16}. Most previously published models train to minimise the negative log-likelihood of the training data, and then at generation time either perform beam search to find the output $Y$ which maximises $P(Y|\mathrm{input})$ \citep{Shang:15,Vinyals:15,Serban:16} (ML decoding), or sample from the resulting distribution \citep{Serban:16}. 

A notorious issue with ML decoding is that this tends to generate short, boring responses to a wide range of inputs, such as \textit{``I don't know"}. These responses are common in the training data, and can be replies to a wide range of inputs \citep{Jiwei:16,Serban:16}. In addition, shorter responses typically have higher likelihoods, and so wide beam sizes often result in very short responses \citep{Tu:2017,Belz:2007}. To resolve this problem, \citet{Jiwei:16} propose instead using maximum mutual information with a length boost as a decoding objective, and report more interesting generated responses.

Further, natural dialogue is not deterministic; for example, the replies to \textit{``What's your name and where do you come from?''} will vary from person to person. \citet{Jiwei:16b} have proposed learning representations of personas to account for inter-person variation, but there can be variation even among a single person's responses to certain questions.

Recently, \citet{Serban:17} have introduced latent variables to the dialogue modelling framework, to model the underlying distribution over possible responses directly. These models have the benefit that, at generation time, we can sample a response from the distribution by first sampling an assignment of the latent variables, and then decoding deterministically. In this way, we introduce stochasticity without resorting to sampling from the decoder, which can lead to incoherent output -- see Table \ref{tab:outputs} for examples.

In this paper, we present a latent variable model for one-shot dialogue response, and investigate what kinds of diversity the latent variables capture. Our experiments show that our model has higher lexical as well as sentential diversity than baseline models. We also show that our model generates more acceptable diverse output than sampling from a deterministic decoder. We end by noting that all three methods proposed above to combat the `maximum likelihood response' are ways of decreasing the probability of the generated output, and report some preliminary results for how response probability interacts with grammaticality and interestingness.

\section{A Latent Variable Dialogue Model}

\begin{table*}[t]
\centering{
\resizebox{\textwidth}{!}{
\begin{tabular}{p{0.3\textwidth} | l l}
\toprule
    Prompt & \textsc{Dial-LV} & Sampling ($\tau=1.0$) \\
\midrule
    \multirow{3}{*}{\pbox{0.3\textwidth}{the infection's spreading.}} 
        & he's dead. & he was lee . \\
        & nothing but no more. & leave it a way. \\
        & i know what you're doing & there's something going on in the phone rickshaw and put the afloat continuously... \\
\midrule
    \multirow{3}{*}{\pbox{0.3\textwidth}{we only have fourteen hours to save the earth!}}
        & i've got no choice . & can i win? \\
        & and then there's something that's what it's like. & go peacekeeper go. \\
        & well, we need the help of what's happening. & go to say hello again in this conversation. \\
\midrule
    \multirow{3}{*}{\pbox{0.3\textwidth}{why don't i do some research and send you an email?}}
        & well, we're gonna have to make it. & no disrespect team. \\
        & because i know how much it is. & excuse me. \\
        & because it 's not right. & because i'm hotel. \\
\bottomrule
\end{tabular}
}
}
\caption{3 random outputs for 3 random prompts from the dataset from our proposed model (\textsc{Dial-LV}) and naively sampling from the decoder of a deterministic encoder-decoder.}
\label{tab:outputs}
\end{table*}

\subsection{Model Description}
Our task is to model the true probability of a response $Y$ given an input $X$. We denote our model distribution by $P(Y | X)$. We introduce a latent variable $z$ with a standard Gaussian prior  -- i.e. $P(z) = \mathcal{N}(0, I_n)$ -- and factor $P(Y | X)$ as:
\begin{equation}
    P(Y | X) = \int_{z} P(Y | z, X) P (z) dz
\label{eqn:chain_rule}
\end{equation}

To motivate this model, we point out that existing encoder-decoder models encode an input $X$ as a single fixed representation. Hence, all of the possible replies to $X$ must be stored within the decoder's probability distribution $P(Y|X)$, and during decoding it is hard to disentangle these possible replies.

However, our model contains a stochastic component $z$ in the decoder $P(Y | z, X)$, and so by sampling different $z$ and then performing ML decoding on $P(Y | z, X)$, we hope to tease apart the replies stored in the probability distribution $P(Y|X)$, without resorting to sampling from the decoder. This has the benefit that we use the decoder at generation time in a similar way to how we train it, making it more likely that the output of our model is grammatical and coherent. Further, as we do not marginalize out $z$ when decoding, we no longer perform exact maximum likelihood search for a reply $Y$, and so we hope to avoid the boring reply problem.

At training time, we follow the variational autoencoder framework \citep{Kingma:14,Kingma:14b,Sohn:2015,Miao:16a} , and approximate the posterior $P(z | X, Y)$ with a proposal distribution $Q(z | X, Y)$, which in our case is a diagonal Gaussian whose parameters depend on $X$ and $Y$. We thus have the following evidence lower bound (ELBO) for the log-likelihood of the data:
\begin{multline}
    \log P(Y | X) \geq -\mathcal{KL} (Q(z | X, Y) || P(z)) \\ + \mathbb{E}_{z \sim Q} \log P(Y | z, X)
\end{multline}

Note that this loss decomposes into two parts: the KL divergence between the approximate posterior and the prior, and the cross-entropy loss between the model distribution and the data distribution. If the model can encode useful information into $z$, then the KL divergence term will be non-zero \citep{Bowman:16}. As our model decoder is given a deterministic representation of $X$ already, $z$ will then encode information about the variation in replies to $X$.

\subsection{Model Implementation}

\begin{figure}[t]
    \centering
    \includegraphics[width=\columnwidth]{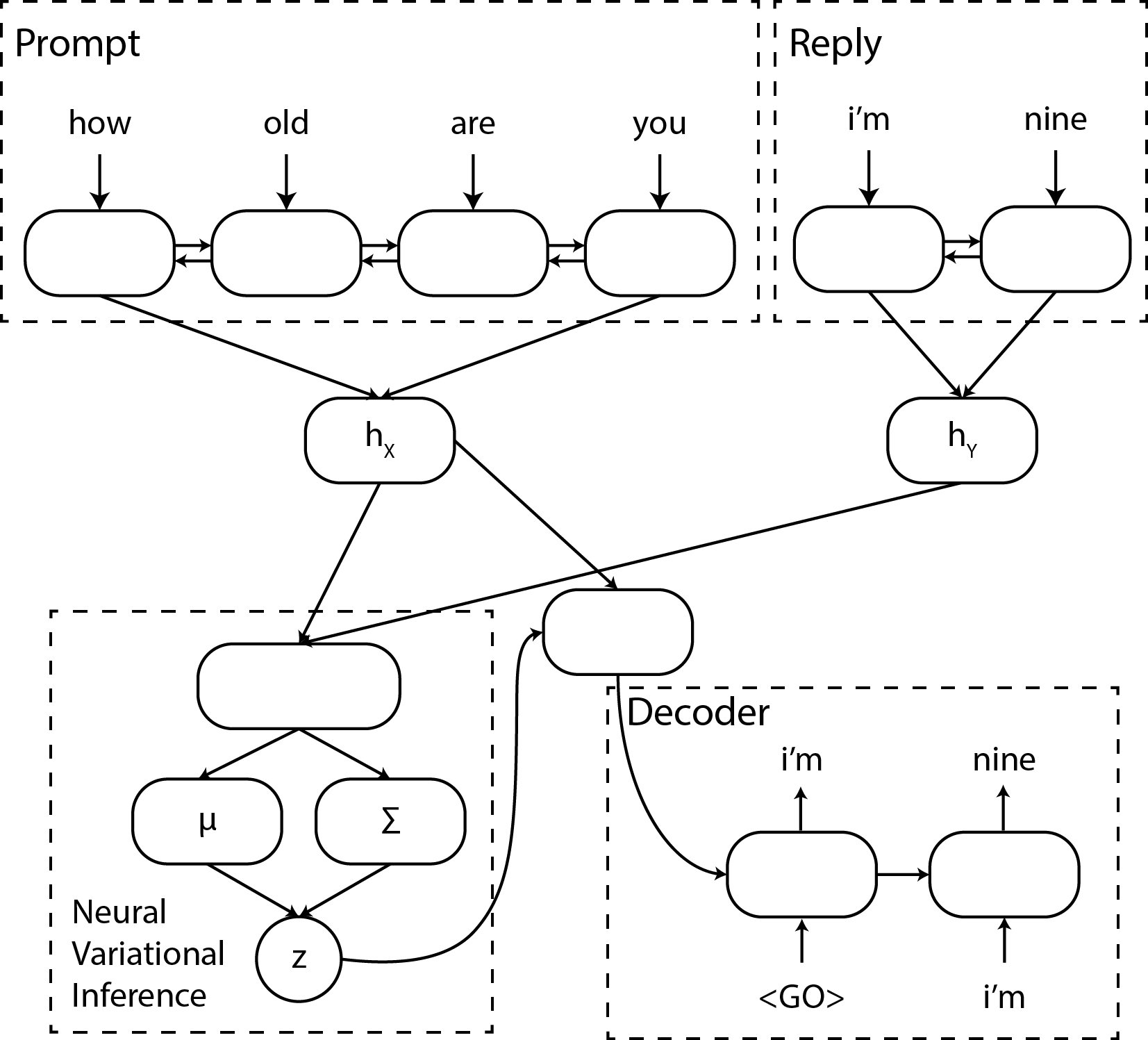}
    \caption{A schematic of how our model is implemented. Please see the text for full details.}
    \label{fig:model_implementation}
\end{figure}

Given an input sentence $X$ and a response $Y$, we run two separate bidirectional RNNs over their word embeddings $\mathbf{x}_i$ and $\mathbf{y}_i$. We concatenate the final states of each and pass them through a single nonlinear layer to obtain our representations $h_{\mathbf{x}}$ and $h_{\mathbf{y}}$ of $X$ and $Y$. We use GRUs \citep{GRU} as our RNN cell as a compromise between expressive power and computational cost.

We calculate the mean and variance of $Q$ as:
\begin{equation}
\begin{split}
   \mu & = W_{\mu} [h_{\mathbf{x}} \ h_{\mathbf{y}}] + b_{\mu} \\
   \log(\Sigma) & = diag(W_{\Sigma} [h_{\mathbf{x}} \ h_{\mathbf{y}}] + b_{\Sigma})
\end{split}
\end{equation}
where $[a \ b]$ denotes the concatenation of $a$ and $b$, and $diag$ denotes inserting along the diagonal of a matrix.

We take a single sample $z$ from $Q$ using the reparametrization trick \citep{Kingma:14}, concatenate $h_{\mathbf{x}}$ and $z$, and initialize the hidden state of the decoder GRU with $[h_{\mathbf{x}} \ z]$. We then train the decoder GRU to minimize the negative log-likelihood of the response $Y$.

While training this model, we noted the same difficulties as \citet{Bowman:16} -- as RNNs are powerful density estimators, the model will prefer to ignore the latent variables and instead optimize the data reconstruction term of the ELBO, while forcing the KL term to 0. We overcome this using similar techniques by gradually annealing the KL term weight over the course of model training and using word dropout in the decoder with a drop rate of $0.5$.

\section{Experiments}

We compare our model, \textsc{Dial-LV}, to three baselines. The first is an encoder-decoder dialogue model with ML decoding (\textsc{Dial-MLE}). The second baseline model implements the anti-LM decoder of \citet{Jiwei:16} (\textsc{Dial-MMI}) on top of the encoder-decoder, with no length normalization. For these models, we use beam search with a width of 2 to find the sentence $Y$ which maximises the decoding objective (either ML or MMI).

The final baseline uses the encoder-decoder model, but instead samples from the decoder to find $Y$ (\textsc{Dial-Samp}). We found that naively sampling from the decoder resulted in meaningless jumbles of words. To solve this, we introduced a temperature parameter $\tau \in (0, 1]$, which scales the probability of each word of the decoder as $p_w \mapsto p_w^{1/\tau}$. This parameter serves to sharpen the word distribution of the decoder. We found $\tau = 0.35$ to be a reasonable balance between preserving stochasticity while also improving the coherence of the generated output.

We used the OpenSubtitles dataset of movie subtitles to train our models \citep{Tiedeman:12}. We took a random sample of 100K files from the full dataset to train our models on, and then pruned this of repeated files to leave roughly 95K files and capped sentence length to 50. The total size of the resulting corpus was around 731M tokens. Please see the supplementary material for model hyperparameters and training details.

As seeds for our replies, we used a list of 50 prompts: 150 lines from the OpenSubtitles dataset outside of our training set which we judged to make sense as independent sentences and 50 questions chosen from a list of suggested conversation starters\footnote{Obtained from http://conversationstartersworld.com/250-conversation-starters/}.

\subsection{Reply statistics}

\begin{table}[t]
\centering{
\resizebox{\columnwidth}{!}{
\begin{tabular}{c|cccc}
\toprule
     Model & Zipf parameter & NLL & Unique \% \\
\midrule
     \textsc{Dial-Lv} & 1.39 & 15.54 & 76 \\
     \textsc{Dial-MLE} & 1.43 & 12.15 & 35 \\
     \textsc{Dial-MMI} & 1.60 & 15.12 & 62 \\
     \textsc{Dial-Samp} & 1.53 & 16.66 & 78 \\
\bottomrule
\end{tabular}
}
}
\caption{Some statistics pertaining to the responses generated by the models.}
\label{tab:statistics}
\end{table}

Previous work (e.g. \citet{Jiwei:16}) used type-token ratio (TTR) to measure the diversity of the generated output. However, as language follows a Zipf distribution, TTR is affected by the length of the generated replies \citep{Mitchell:2015}. Hence, we use the estimated parameter of a Zipf distribution fitted to our replies as a proxy for the lexical diversity of generated output, with more diverse output having smaller scores. As ML decoding is known to give the same few replies repeatedly, we also report the percentage of unique replies, as a coarser measure of sentential diversity compared to lexical diversity. Further, we give the negative log-likelihood (NLL) as predicted by the deterministic encoder-decoder model, to see what regions of the probability space the replies occupy. We present these statistics in Table \ref{tab:statistics}. 

We note that \textsc{Dial-LV} generates more diverse replies than the other deterministic models, measured in terms of percentage of unique responses. Interestingly, the lexical diversity of \textsc{Dial-LV} is almost identical to \textsc{Dial-MLE}, suggesting that the latent variables help \textsc{Dial-LV} avoid the boring output problem and generate more diverse outputs. We note that \textsc{Dial-LV} even rivals \textsc{Dial-Samp} in terms of sentential diversity, and beats \textsc{Dial-Samp} in terms of lexical diversity. This could be because \textsc{Dial-Samp} chooses words greedily, and so is biased towards choosing high-probability words at each timestep. This suggests that maintaining a beam of hypotheses while sampling could help sampling-based methods escape the trap of having to make near-greedy local decisions.

\subsection{Human acceptability judgments}

\begin{table}[t]
\centering{
\resizebox{\columnwidth}{!}{
\begin{tabular}{c|ccccc}
\toprule
     Model & $\mu$ & $\sigma$ & NLL & Zipf & Unique \% \\
\midrule
     \textsc{Dial-LV} & 1.183 & 0.402 & 15.51 & 1.32 & 76.4 \\
     \textsc{Dial-Samp} & 1.196 & 0.577 & 16.91 & 1.56 & 73.6 \\
\bottomrule
\end{tabular}
}
}
\caption{Mean and std. dev. of average number of acceptable replies generated by each model.}
\label{tab:acceptability}
\end{table}

We also tested whether \textsc{Dial-LV} could generate a greater number of acceptable replies to a prompt than \textsc{Dial-Samp}. We randomly selected 50 prompts from our list of 200, and generated 5 replies at random to each one using both models. We then asked human annotators\footnote{We used 50 in total, 25 for each model} to judge how many replies were appropriate replies, taking into account grammaticality, coherence and relevance. The results are shown in Table \ref{tab:acceptability}.

Interestingly, even though \textsc{Dial-LV} has a lower NLL score, both models generate roughly the same number of acceptable replies. \textsc{Dial-LV} also has less variance in the number of acceptable replies, suggesting that the outputs it generates are more consistent than responses from \textsc{Dial-Samp}. Finally, we note that \textsc{Dial-LV} generates more diverse output than \textsc{Dial-Samp} in this scenario, even thought its replies are judged equally acceptable, suggesting that it is managing to produce a wide range of coherent, fluent and appropriate output.

\subsection{Sampling from the latent variable space}

\begin{table}[t]
\centering{
\resizebox{\columnwidth}{!}{
\begin{tabular}{c|ccc}
\toprule
     Shell radius & Zipf parameter & NLL & Unique \% \\
\midrule
     0 & 1.49 & 13.12 & 7 \\
     4 & 1.62 & 14.02 & 42.1 \\
     8 & 1.59 & 15.72 & 63.1 \\
     12 & 1.56 & 17.65 & 67.7 \\
     16 & 1.78 & 18.16 & 67.1 \\
\bottomrule
\end{tabular}
}
}
\caption{Statistics of responses generated from the \textsc{Dial-LV} model from different regions of the hidden state space.}
\label{tab:variety}
\end{table}

We next explored the effect of sampling from different regions of the latent space. For each prompt in the test set, we took 5 uniform samples from shells of radius 0 (which collapses to deterministic decoding), 4, 8, 12 and 16 in the latent space\footnote{For a $d$-dim standard Gaussian, $\mathbb{E}(\|X\|) \approx \sqrt{d}$, and $Var(\|X\|) \to 0$ as $d \to \infty$. Here $d=64$.} by sampling from $P(z) = \mathcal{N}(0, I)$ and then scaling the sample $z$ by the appropriate amount. We then generated a response to the prompt using each value of $z$, and measured some statistics of the replies. The results are shown in Table \ref{tab:variety}.

As expected, samples with small radius show less diversity in terms of unique outputs. Further, we see a consistent trend that samples with greater radius have a higher NLL score, showing the influence of the prior in Eqn. \ref{eqn:chain_rule}. However, at the highest radius, we observe the highest NLLs, but also the lowest lexical diversities, suggesting that it manages to combine the words it produces in many different ways.

\section{Discussion}
\label{sec:discussion}
Taken together, our experiments show that ML decoding does not seem to be the best objective for generating diverse dialogue, and so corroborates the inadequacy of perplexity as an evaluation metric for dialogue models \citep{Liu:2016}. Indeed, all three models which show a diversity gain over the vanilla encoder-decoder with MLE decoding try to instead sample responses from a lower-probability region of the response space. However, if the response probability is too low, it runs the risk of being nonsensical. Hence, there appears to be a `Goldilocks' region of the probability space, where the responses are interesting and coherent. Finding ways of concentrating model samples to this region is thus a potentially promising area of research for open-domain dialogue agents.

We also note that our proposed model can be combined with MMI decoding or temperature-based sampling to get the benefits of both worlds. While we did not do this in our experiments in order to isolate the impact of our model, doing so improves the diversity of our generated output even more.

\section{Conclusion}
In this paper, we present a latent variable model to generate responses to input utterances. We investigate the diversity of output generated from this model, and show that it improves both lexical and sentential diversity. It also generates more consistently acceptable output as judged by humans compared to sampling from a decoder.

\section*{Acknowledgements}

KC is supported by an EPSRC doctoral award. SC is supported by ERC Starting Grant DisCoTex (306920) and ERC Proof of Concept Grant GroundForce (693579). The authors would like to thank everyone who helped prototype the human evaluation experiments. The authors would also like to thank the anonymous reviewers for all their insightful comments.

\bibliography{eacl2017}
\bibliographystyle{plainnat}

\appendix

\section{Model training information}
\label{sec:supplemental}
We implemented all of our models using Keras \citep{keras} running on Theano \citep{theano1}. As vocabulary, we took all words appearing at least 1000 times in the whole corpus. As this amounted to $\sim$30K words, we used a 2-level hierarchical approximation to the full softmax to speed up model training \citep{Morin:05}, with random clustering. We trained all our models for 3 epochs using the Adadelta optimizer \citep{Zeiler:12}, with default values for the optimizer parameters.

We used 512 dimensional word embeddings and encoder hidden state sizes across all of our models. We used 64 latent dimensional latent variables, and so the decoder RNN for the \textsc{Dial-LV} model had hidden state size 576. The decoder RNN for the \textsc{Dial-MLE} model also had hidden state size 576, to keep the capacity of the decoder comparable across the two models. We used tanh non-linearities throughout our model. For training the vanilla encoder-decoder, we also used word dropout on the decoder input with a drop rate of $0.5$ to prevent overfitting. Each epoch took roughly 4 days on a Titan Black.

For the MMI decoding, we used a LM penalty weight of $0.45$ and applied this for the first $6$ words.

\end{document}